\documentclass[10pt,twocolumn,letterpaper]{article}

\usepackage{wacv}              

\usepackage{graphicx}
\usepackage{amsmath}
\usepackage{amssymb}
\usepackage{booktabs}
\usepackage{times}
\usepackage{epsfig}
\usepackage{dsfont}
\usepackage{multirow}
\usepackage{xcolor}
\usepackage{paralist}

\usepackage[pagebackref,breaklinks,colorlinks]{hyperref}

\usepackage[capitalize]{cleveref}
\crefname{section}{Sec.}{Secs.}
\Crefname{section}{Section}{Sections}
\Crefname{table}{Table}{Tables}
\crefname{table}{Tab.}{Tabs.}

\def\wacvPaperID{1222} 
\def\confName{WACV}
\def\confYear{2024}

\begin{document}

\title{
Hierarchical Text Spotter for Joint Text Spotting and Layout Analysis}

\author{Shangbang Long, Siyang Qin, Yasuhisa Fujii, Alessandro Bissacco, Michalis Raptis\\
Google Research\\
{\tt\small \{longshangbang,qinb,yasuhisaf,bissacco,mraptis\}@google.com}}

\maketitle

\begin{abstract}
   We propose \textbf{Hierarchical Text Spotter (HTS)}, a novel method for the joint task of word-level text spotting and geometric layout analysis.
   HTS can recognize text in an image and identify its $4$-level hierarchical structure: characters, words, lines, and paragraphs.
   The proposed HTS is characterized by two novel components: 
   (1) a \textbf{Unified-Detector-Polygon (UDP)} that produces Bezier Curve polygons of text lines and an affinity matrix for paragraph grouping between detected lines; 
   (2) a \textbf{Line-to-Character-to-Word (L2C2W)} recognizer that splits lines into characters and further merges them back into words.
   HTS achieves state-of-the-art results on multiple word-level text spotting benchmark datasets as well as geometric layout analysis tasks.
\end{abstract}

\section{Introduction}
\label{sec:intro}

\begin{figure}[t]
  \includegraphics[width=\linewidth]{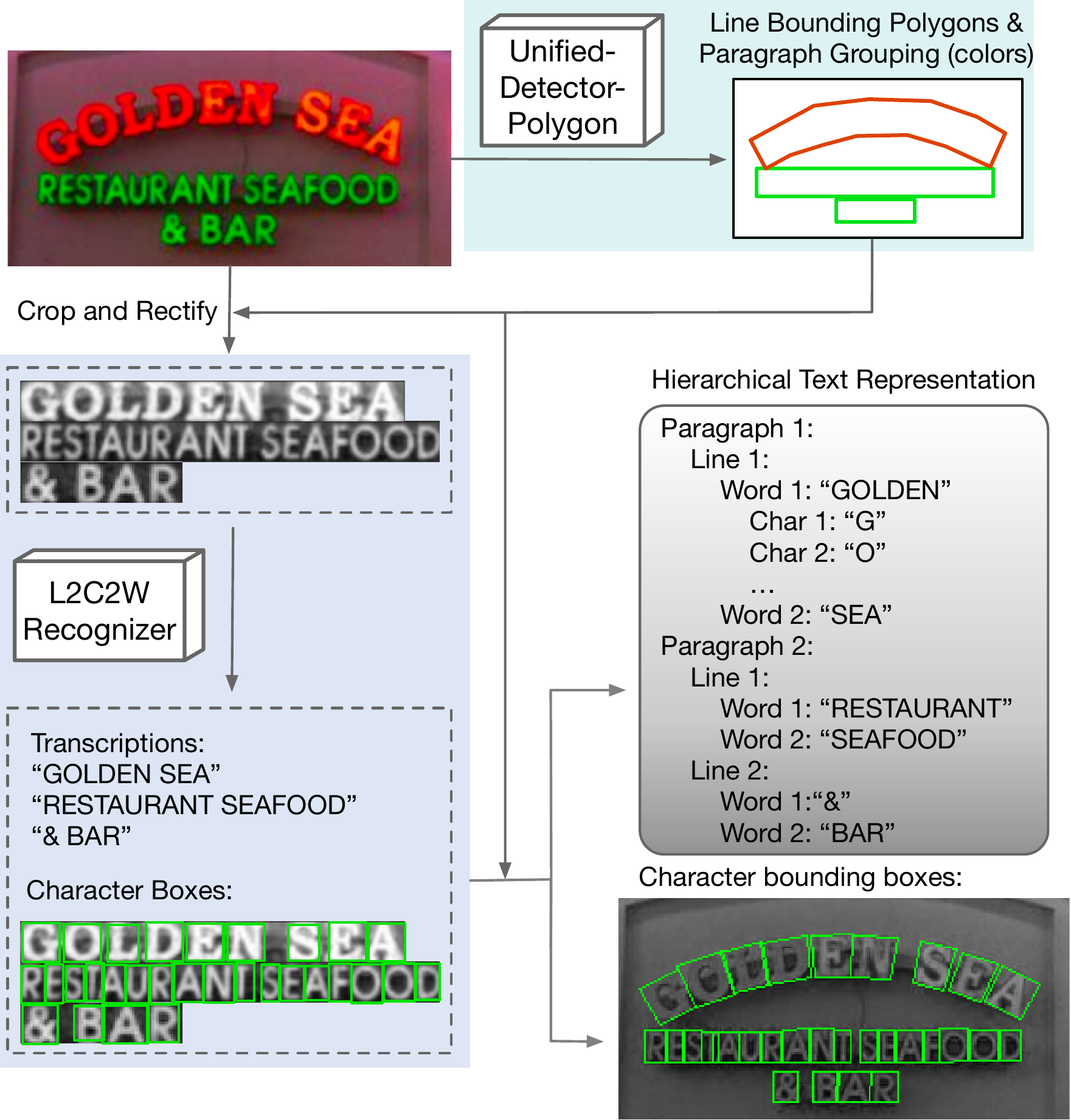}
  \caption{Illustration of our \textbf{Hierarchical Text Spotter (HTS)}. 
  HTS consists of two main components: 
  (1) \textbf{Unified-Detector-Polygon (UDP)} that detects text lines with bounding polygons and clusters them into paragraph groups. 
  In this figure \textit{(upper right)}, paragraph groups are illustrated by different colors.
  The bounding polygons are used to crop and rectify text lines into canonical forms that are easy to recognize.
  (2) \textbf{Line-to-Character-to-Word (L2C2W)} Recognizer that jointly predicts character classes and bounding boxes.
  Spaces are used to split lines into words.
  The output of HTS is a \textbf{Hierarchical Text Representation (HTR)} that encodes the layout of all text entities in an image.
  In this figure, we use indents to represent the hierarchy of text entities \textit{(middle right)}, and visualize the character bounding boxes \textit{(bottom right)}.
  }
  \vspace{-3mm}
\label{fig:framework}
\end{figure}

The extraction and comprehension of text in images play a critical role in many computer vision applications. Text spotting algorithms have progressed significantly in recent years \cite{qin2019towards,long2021scene,zhang2022text,ronen2022glass,peng2022spts}, specifically within the task of detecting \cite{long2018textsnake,baek2019character,liao2022real,ye2022dptext} and recognizing \cite{wang2022multi,da2022levenshtein,bautista2022scene,na2022multi,nuriel2022textadain} individual text instances in images.
Previously, defining the geometric layout \cite{cattoni1998geometric,breuel2002two,yang2017learning,lee2019page} of extracted textual content occurred independent of text spotting and remained focused on document images. In this paper, we aim to further the argument \cite{long2022towards} that consolidating these separately treated tasks is complementary and mutually enhancing.
We postulate a joint approach for text spotting and geometric layout analysis could provide useful signals for downstream tasks such as semantic parsing and reasoning of text in images such as text-based VQA  \cite{biten2019scene,singh2019towards}
and document understanding \cite{li2021structurallm,lee2022formnet,huang2022layoutlmv3}.

Existing text spotting methods \cite{qin2019towards,zhang2022text,ronen2022glass} most commonly extract text at the word level, where `word' is defined as a sequence of characters delimited by space without taking into account the text context.
Recently, the Unified Detector \cite{long2022towards}, which is built upon detection transformer \cite{wang2021max}, detects text `lines' with instance segmentation mask and produces an affinity matrix for paragraph grouping in an end-to-end way. This method is limited to the detection task and it can not produce character or word-level outputs.

In this paper, we propose a novel method, termed \textit{Hierarchical Text Spotter (HTS)}, that simultaneously localize, recognize and recovers the geometric relationship of the text on an image.
The framework of HTS is illustrated in Fig. \ref{fig:framework}.
It is designed to extract a hierarchical text representation (HTR) of text entities in images.
HTR has four levels of hierarchy\footnote{Here, we follow the definitions of these levels in \cite{long2022towards}.}, including \textit{character}, \textit{word}, \textit{text line}, and \textit{paragraph}, from bottom to top. 
The HTR representation encodes the structure of text in images.
To the best of our knowledge, HTS is the first unified method for text spotting and geometric layout analysis.

The proposed HTS consists of two main components: 
(1) A \textit{Unified-Detector-Polygon (UDP)} model that jointly predicts Bezier Curve polygons \cite{liu2021abcnet} for text lines and an affinity matrix supporting the  grouping of lines to paragraphs. 
Notably, we find that the conventional way of training Bezier Curve polygon prediction head, i.e. applying $L1$ losses on control points directly \cite{liu2021abcnet,raisi2022arbitrary,tang2022few}, fails to capture text shapes accurately on highly diverse dataset such as HierText \cite{long2022towards}.
Hence, we propose a novel \textit{Location and Shape Decoupling Module (LSDM)} which decouples the representation learning of location and shape.
UDP equipped with LSDM can accurately detect text lines of arbitrary shapes, sizes and locations across multiple datasets of different domains.
(2) A \textit{Line-to-Character-to-Word (L2C2W)} text line recognizer based on Transformer encoder-decoder \cite{vaswani2017attention} that jointly predicts character bounding boxes and character classes. 
L2C2W is trained to produce the special space character to delimit text lines into words.
Also, unlike other recognizers or text spotters that are based on character detection \cite{xing2019convolutional,liao2019scene,baek2020character,long2020new}, L2C2W only needs a small fraction of training data to have bounding box annotations.

The proposed HTS method achieves state-of-the-art text spotting results on multiple datasets across different domains, including ICDAR 2015 \cite{karatzas2015icdar}, Total-Text \cite{ch2017total}, and HierText \cite{long2022towards}. 
It also surpasses Unified Detector \cite{long2022towards} on the geometric layout analysis benchmark of HierText, achieving new state-of-the-art result.
Importantly, these results are obtained with a single model, without fine-tuning on target datasets; ensuring that the proposed method can support generic text extraction applications. In ablation studies, we also examine our key design choices.

Our core contributions can be summarized as follows:

\begin{compactitem}
    \item A novel Hierarchical Text Spotter for the joint task of word-level text spotting and geometric layout analysis.
    \item Location and Shape Decoupling Module which enables accurate polygon prediction of text lines on diverse datasets.
    \item L2C2W that reformulates the role of recognizer in text spotter algorithms by performing part of layout analysis and text entities localization.
    \item State-of-the-art results on both text spotting and geometric layout analysis benchmarks without fine-tuning to each particular test dataset.
\end{compactitem}

\section{Related Works}
\noindent \textbf{Text Spotting} Two-stage text spotters consist of a text detection stage and a text recognition stage.
Text detection stage produces bounding polygons or rotated bounding boxes for text instances at one granularity, usually words. 
Text instances are cropped from input image pixels \cite{bartz2018see}, encoded backbone features \cite{qin2019towards,liao2020mask}, or both \cite{ronen2022glass}.
The text recognition stage decodes the text transcription.
End-to-end text spotters use feature maps for the cropping process.
In this case, the text recognition stage reuses those features, improving the computational efficiency \cite{liu2018fots}.
However, end-to-end text spotters suffer from asynchronous convergence between the detection and the recognition branch \cite{krylov2021open}.
Due to this challenge, our proposed HTS crops from input image pixels with bounding polygons.
The aforementioned text spotter framework connects detection and recognition explicitly with detection boxes.
Another branch of two-stage text spotter performs implicit feature feeding via object queries \cite{kittenplon2022towards} as in detection transformer \cite{carion2020end} or deformable multi-head attention \cite{zhang2022text}.
More recently, single stage text spotters \cite{peng2022spts,kil2023towards} are proposed under a sequence-to-sequence framework.
These works do not perform layout analysis and are thus orthogonal to this paper.

\noindent \textbf{Text Detection} 
Top-down text detection methods view text instances as objects. These methods produce detection boxes \cite{liu2021abcnet,tang2022few} or instance segmentation masks \cite{long2022towards} for each text instance.
Bottom-up methods first detect sub-parts of text instances and then connect these parts to construct whole-text bounding boxes \cite{shi2017detecting} or masks \cite{long2018textsnake}.
Top-down methods tend to have simpler pipelines, while bottom-up techniques excel at detecting text of arbitrary shapes and aspect ratios.
Neither top-down nor bottom-up mask prediction methods are proficient for spotting curved text, because a mask can only locate text but cannot rectify it.
Additionally, the performance of such models on curved text datasets is commonly reported by fine-tuning those models on the specific data. Therefore, it is unknown whether polygon prediction methods can adapt to text of arbitrary shapes and aspect ratios on diverse datasets.

\noindent \textbf{Text Recognition} An important branch of text recognizers \cite{long2019rethinking, da2022levenshtein,na2022multi} formulates the task as a sequence-to-sequence task \cite{sutskever2014sequence}, where the only output target is a sequence of characters. Another branch formulates the task as character detection \cite{liao2019scene,long2020new}, where it produces character classes and locations simultaneously. However, it requires bounding box annotations on all training data, which are rare for real-image data. Our recognition method falls into the sequence-to-sequence learning paradigm, with the additional ability to produce each character bounding box. Importantly, our model's training requires only partially annotated data, i.e. only a fraction of the data needs to include character level bounding box annotations.

\noindent \textbf{Layout analysis}
Geometric layout analysis \cite{zhong2019publaynet,pfitzmann2022doclaynet,jaume2019funsd,wang2022post} aims to detect visually and geometrically coherent text blocks as objects.
Recent works formulate this task as object detection \cite{schreiber2017deepdesrt}, semantic segmentation \cite{lee2019page,long2022towards}, or learning on the graphical structure of OCR tokens via GCN \cite{wang2022post}. 
Almost all entries in the HierText competition at IDCAR 2023 \cite{hiertext_comp_2023} adopt the segmentation formulation.
Unified Detector \cite{long2022towards} consolidates the task of text line detection and geometric layout analysis. However, it can not produce word-level entities and does not provide a recognition output.
Another line of layout analysis research focuses on semantic parsing of documents \cite{li2021structurallm,lee2022formnet,huang2022layoutlmv3} to identify key-value pairs.
These methods build language models \cite{radford2018improving,devlin2018bert} on top of OCR results.
Recently, StruturalLM \cite{li2021structurallm} and LayoutLMv3 \cite{huang2022layoutlmv3} show that the grouping of words into \textit{segments} using heuristics, which is equivalent to text line formation, improves parsing results.
We believe our work of jointly text spotting and geometric layout analysis can benefit semantic parsing and layout analysis.

\section{Methodology}
\subsection{Hierarchical Text Spotter}
As illustrated in Fig.~\ref{fig:framework} our HTS method mainly comprises $2$ stages:
(1) \textit{Unified Detection Stage}: we propose an end-to-end trainable model termed {Unified-Detector-Polygon (UDP)} that detects text lines in the form on Bezier Curve Polygons \cite{liu2021abcnet}, and simultaneously clusters them into paragraphs.
UDP contains the Location and Shape Decoupling Module (LSDM), a key component in accurate text line detection across diverse datasets.
Text line images are cropped from the input image with BezierAlign \cite{liu2021abcnet} and then converted to grayscale image patches.
(2) \textit{Line Recognition Stage}: We propose an autoregressive text line recognizer based on Transformer encoder-decoder \cite{vaswani2017attention} that jointly predicts character bounding boxes and character classes. 
We train our recognizer to identify printable characters and a special non-printable \textit{space} delimiter.
We use the \textit{space} character to split text lines into word-level granularity. The word-level bounding boxes are formed from the predicted character-level bounding boxes.
Character and word bounding boxes are estimated in the coordinate space of text line image patches. During the post-processing step, they are projected back to the input image coordinate space.
Putting these together, we obtain a hierarchical text representation of \textit{character}, \textit{word}, \textit{line}, and \textit{paragraph}.

\subsection{Unified Detection of Text Line and Paragraph}

\begin{figure}[t]
\begin{center}
  \includegraphics[width=\linewidth]{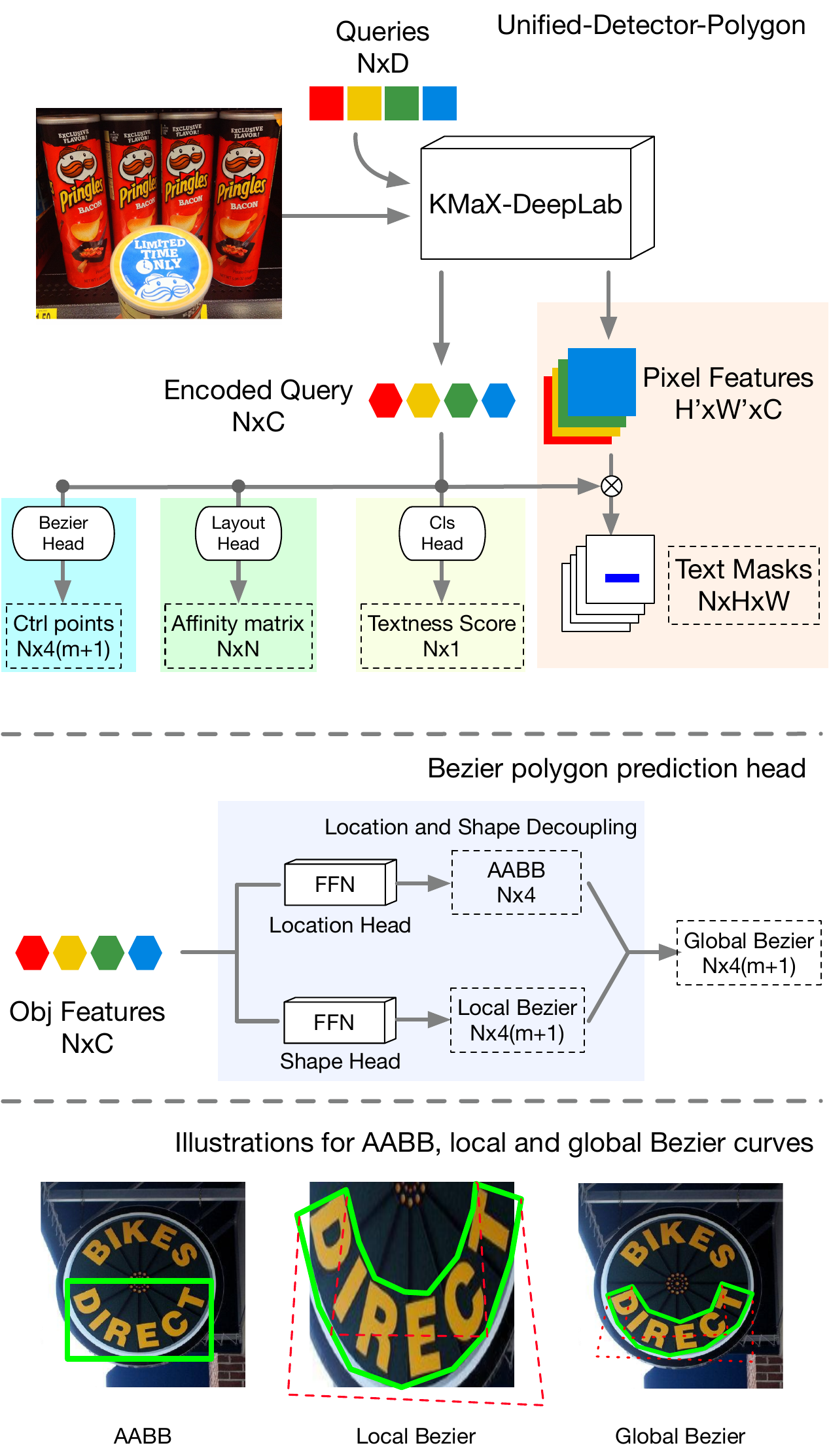}
\end{center}
  \caption{Illustration of our \textbf{Unified-Detector-Polygon (UDP)}. 
  \textbf{Top}: Architecture of UDP, where each color tint represents one prediction branch.
  $N$ is the number of queries. $m$ is the order of the Bezier Curves. $C$ is the model width. $D$ is the query dimension.
  \textbf{Middle}: Architecture of our Bezier polygon prediction head with a dual-head Location and Shape Decoupling Module.
  \textbf{Bottom}: Illustrations for axis-aligned bounding box (AABB), local and global Bezier curve representation.
}
  \vspace{-3mm}
\label{fig:detector}
\end{figure}

\noindent \textbf{Preliminaries} 
Based on MaX-DeepLab \cite{wang2021max}, Unified-Detector \cite{long2022towards} detects text lines by producing instance segmentation masks from the inner product of object queries and pixel features.
Further, an affinity matrix that represents the paragraph grouping is produced by computing the inner product of layout features which are extracted by extra transformer layers applied on object queries.

\noindent \textbf{Unified-Detector-Polygon (UDP)} While Unified Detector \cite{long2022towards} achieves state-of-the-art text detection performance, it uses only masks to localize text instances. The estimated masks can not be directly used to rectify curved text lines.  Thus, complex post-processing heuristics are required to build an effective text-spotting system.
Therefore, we extend the model with an additional Bezier polygon prediction head applied on the encoded object queries, as illustrated in the top of Fig.~\ref{fig:detector}. The Bezier polygon prediction head produces a polygon representation \cite{liu2021abcnet} based on Bezier Curve\footnote{\url{https://en.wikipedia.org/wiki/Bezier_curve}}.
In this representation, each text line is parametrized as two Bezier Curves of order $m$, one for the top and one for the bottom polyline of the text boundary.
Each Bezier Curve has $m+1$ control points.
The model is trained to predict these $2(m+1)$ control points i.e. $4(m+1)$ coordinates.
During inference, the text boundaries are reconstructed from the predicted control points.
In addition, we also replace MaX-DeepLab in the original Unified Detector with KMaX-DeepLab \cite{yu2022k} as the backbone, which is faster and more accurate.

\noindent \textbf{Location and Shape Decoupling Module}
Previous works \cite{raisi2022arbitrary,tang2022few} use a single feed-forward neural network (FFN) to predict the control points in image space and train the network by applying L1 loss on the control points.
However, as shown in Sec.~\ref{sec:ablation}, such approach results to sub-optimal detection accuracy for text line datasets such as HierText \cite{long2022towards} due to its diverse locations, aspect ratios, and shapes.
To mitigate this issue, we propose a novel \textit{Location and Shape Decoupling Module (LSDM)}.
As shown in the middle of Fig.~\ref{fig:detector}, it consists of two parallel FFNs, one for location prediction and the other for shape prediction.
The Location Head predicts Axis-Aligned Bounding Boxes (AABB) whose coordinates are normalized in the image space.
For the $i-$th text instance, we denote its predicted AABB as:

\vspace{-5mm}

\begin{equation}
    {AABB}_i=[{x}_{\mathrm{center},i}, {y}_{\mathrm{center},i}, {w}_i, {h}_i] \in R^4
    \label{eq:aabb}
\end{equation}

\noindent  representing its center, width, and height.
The Shape Head predicts Local Bezier Curve control points whose coordinates are normalized in the space of the AABB:
\begin{equation}
    {bezier}_{local,i} = \{({\tilde{x}}_{i,j}, {\tilde{y}}_{i, j})\}_{j=1}^{2(m+1)}
    \label{eq:local-bezier}
\end{equation}

Finally, the Global, i.e. image space, Bezier curve control point coordinates are obtained by scaling and translating Local Bezier coordinates by AABB:
\begin{align}
    {bezier}_{global,i} = \{({x}_{i,j}, {y}_{i, j})\}_{j=1}^{2(m+1)}  \label{eq:global-bezier1} \\
     \text{where}\  {x}_{i,j} = {\tilde{x}}_{i, j} * {w}_i + {x}_{\mathrm{center},i} \label{eq:global-bezier2}\\
      {y}_{i,j} = {\tilde{y}}_{i, j} * {h}_i + {y}_{\mathrm{center},i} \label{eq:global-bezier3}
\end{align}

The concepts of AABB, Local Bezier coordinates, and Global Bezier coordinates are further illustrated in the bottom of Fig.~\ref{fig:detector}.
During training, we generate appropriately ground-truth data for both heads and apply supervision on both of them.
Specifically, given ground-truth Global Bezier control points, we first compute ground-truth AABB as the minimum area AABB enclosing the ground-truth polygons, and then use the reverse of Eq. (\ref{eq:global-bezier1}) (\ref{eq:global-bezier2}) (\ref{eq:global-bezier3}) to compute ground-truth Local Bezier control points.
The final training loss is the weighted sum of all Unified Detector \cite{long2022towards} loss, GIoU loss on AABB \cite{rezatofighi2019generalized}, L1 loss on AABB, and L1 loss on local control points:
\vspace{-3mm}
\begin{multline}
    L_{det} = L_{\text{unified detector}} \\ + \lambda_1L_{\text{AABB}, \mathrm{GIoU}} + \lambda_2L_{\text{AABB}, L_{1}}  + \lambda_3L_{\text{Local}, L_{1}}
\end{multline}
\noindent where $\lambda_1$, $\lambda_2$, $\lambda_3$ are the weights for loss balancing.

\subsection{Line-to-Character-to-Word Recognition}
\begin{figure}[t]
\begin{center}
  \includegraphics[width=1.0\linewidth]{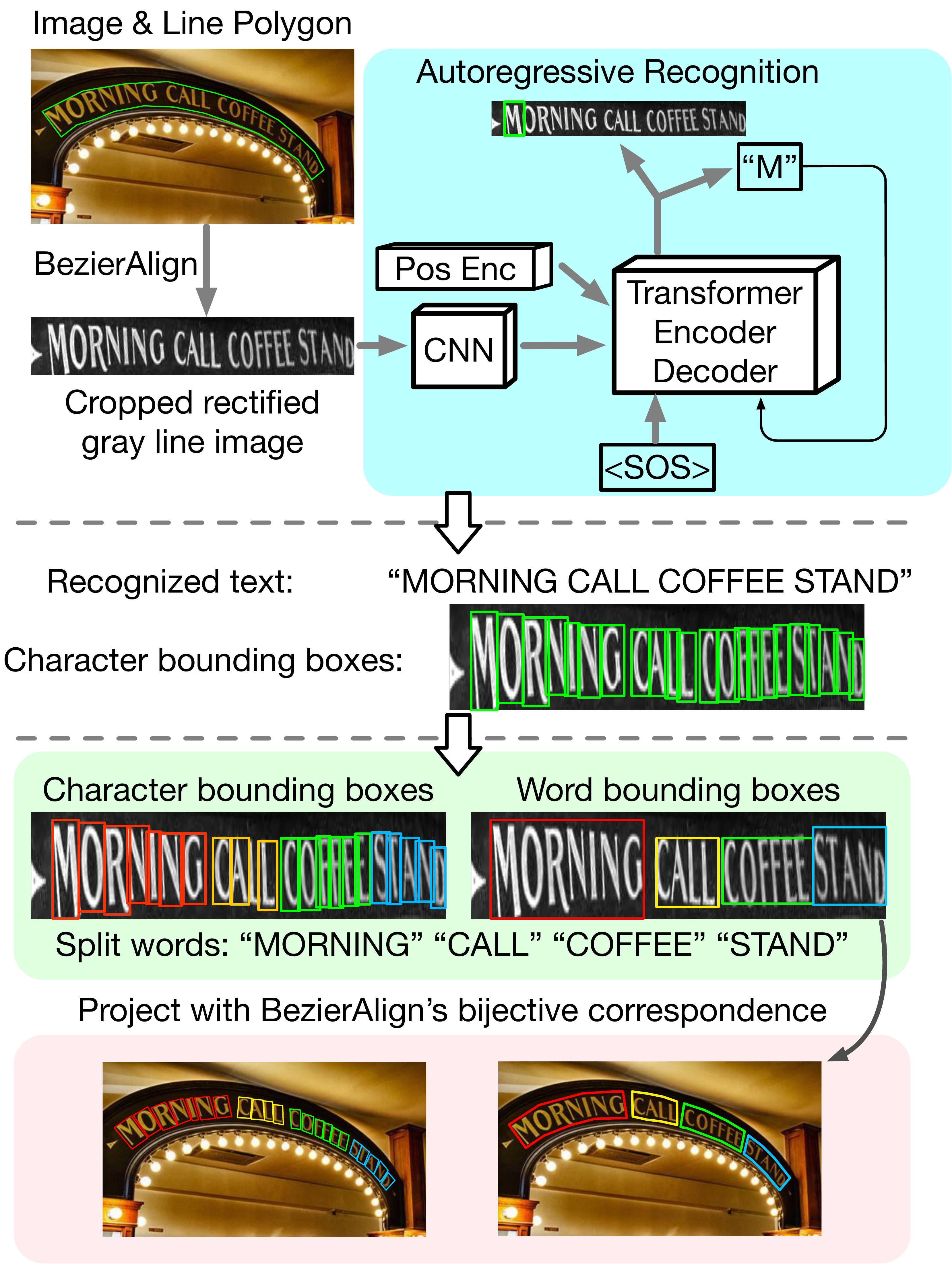}
\end{center}
  \caption{Illustration of our \textbf{Line-to-Character-to-Word (L2C2W)} recognition method.
  \textbf{Top}: Text line images are cropped and rectified from the input image using BezierAlign \cite{liu2021abcnet}.
  Our L2C2W recognition model uses an autoregressive transformer encoder-decoder model \cite{vaswani2017attention} to predict character class and box simultaneously.
 \textbf{Middle}: Output sample.
 \textbf{Bottom}: Text line recognition results are split into words, and character bounding boxes are clustered in accordance with words and form word bounding boxes.
  Bounding boxes are projected back to image space.
  }
  \vspace{-10mm}
\label{fig:rec}
\end{figure}

We propose a novel hierarchical text recognition framework, termed \textit{Line-to-Character-to-Word (L2C2W)}.
 Fig.~\ref{fig:rec} illustrated our framework.
Text line images are cropped and rectified from the input image with BezierAlign \cite{liu2021abcnet}.
We use the grayscale cropped image as input for the recognizer.
The model predicts character-level outputs.
To correctly group characters into words, our recognition model learns to predict both printable characters and the \textit{space} character.
During inference, we use the space as the delimiter to segment a text line string into words. The model also produces character-level bounding boxes. These character bounding boxes are grouped based on each word's boundaries and produce the words' bounding boxes.

\noindent \textbf{Text Line Recognition Model}
Our transformer-based recognizer consists of three stages.
First, a MobileNetV2 \cite{sandler2018mobilenetv2} convolutional backbone encodes the image pixels, and reduces the height dimension to $1$ using strided convolutions.
Then, a sinusoidal positional encoding \cite{vaswani2017attention} is added, and transformer encoder layers are applied on the encoded features.
Lastly,  a transformer decoder produces the predicted output autoregressively \cite{sutskever2014sequence}.

\noindent \textbf{Character Localization}
We use axis-aligned bounding boxes to represent the location of characters in cropped text lines.
Vanilla transformer decoder \cite{vaswani2017attention} has only one prediction head to produce a probability distribution over the next token.
To predict character bounding boxes, we add a 2-layer FFN prediction head on the output feature from decoder, in parallel to the classification head.
The character location head produces a $4$d vector representing the top-left and bottom-right coordinates of the character bounding boxes.
These character coordinates are normalized by each text line's height.

\noindent \textbf{Training} The total loss for training is the weighted sum of character classification loss and character localization loss.
We use cross-entropy for character classification and $L_1$ loss for character localization.
It is important to note that ground-truth annotated character bounding boxes are rare in real-image datasets but are available in most synthetic text data \cite{gupta2016synthetic,long2020unrealtext,yim2021synthtiger}.
During training, we mix real-image and synthetic data and apply character localization loss only when ground-truth labels are available.
The training target for one text line can be formulated as:
\vspace{-3mm}

\begin{multline}
    L_{rec} = \frac{1}{T}\sum_{t=1}^{T} L_{CE}(y_t, \hat{y}_t)
    + \frac{\lambda_4 \sum_{t=1}^{T} \alpha_t L_{\mathrm{L}_1}(\mathrm{box}_t, \widehat{\mathrm{box}}_t)}{\sum_{t=1}^{T} \alpha_t + \epsilon}
\end{multline}
\noindent where $T$ is the number of characters, $\lambda_4$ is the weight for localization loss, $\alpha_t$ is an indicator for whether it has ground-truth character bounding box, and $\epsilon$ is a small positive number to avoid zero denominator.
In practice, we do the summation and average on a batch level, to balance the loss between long and short text.

\noindent \textbf{Post-processing}
We partition text lines into words using the predicted \textit{space} character. We obtain word-level bounding boxes by finding the minimum-area axis-aligned bounding boxes of each word's characters. Finally, we project these word and character bounding boxes back to the image space.
When we perform BezierAlign \cite{liu2021abcnet} in line cropping, we build a bijection from coordinates in text line crops to coordinates in the input image. 
We re-use this bijection to compute this projection operation (detailed in \textit{supplementary material Sect. A}).

\section{Experiments}
In this section, we evaluate the proposed method on a number of benchmarks.
We first introduce the experimental settings, including the training and test datasets, the hyper-parameters of models, and the evaluation practices. We compare our method to the current state-of-the-art on end-to-end text spotting, text detection, and geometric layout analysis \cite{long2022towards}.
Finally, we conduct comprehensive ablation studies and analyze our design choices.

\subsection{Experiment Setting}

\noindent\textbf{Unified-Detector-Polygon}
We base our UDP implementation on the official repository\footnote{\url{https://github.com/tensorflow/models/tree/master/official/projects/unified_detector}} of Unified Detector \cite{long2022towards}.
The input resolution is $1600\times1600$.
Model dimensions are $N=384$, $D=256$, $C=128$ respectively, using the same settings as Unified Detector \cite{long2022towards}.
As for the Bezier polygon prediction head, we use a $2$-layer MLPs for both branches, with a hidden state dimension of $256$.
ReLU and LayerNorm \cite{ba2016layer} are applied in between the two layers.
The AABB and Local Bezier prediction head outputs are activated by a sigmoid and a linear function respectively.
We use $m=3$, i.e. cubic order Bezier Curves.
For the loss balancing weights, we set $\lambda_1=1.0$, $\lambda_2=2.5$, and $\lambda_3=0.5$.
The ratio of $\lambda_1$ and $\lambda_2$ are set after DETR \cite{carion2020end}.
UDP is trained on $128$ TPUv3 devices for $100K$ iterations with a batch size of $256$, AdamW \cite{loshchilov2017decoupled} optimizer, cosine learning rate \cite{loshchilov2016sgdr} of $0.001$, and weight decay of $0.05$.
We train UDP on a combination of the training sets of HierText \cite{long2022towards} and CTW1500 \cite{yuliang2017detecting}, which both provide line-level text annotations.
During training, images are randomly rotated, cropped, padded, and resized to the input resolution.
A random scheme of color distortion \cite{cubuk2020randaugment} is also applied.

\noindent \textbf{L2C2W Recognizer}
We use the TensorFlow Model Garden library \cite{tensorflowmodelgarden2020} to implement our model.
Input text lines are resized to height of $40$ pixels and padded to width of $1024$ pixels to accommodate the variable aspect ratios of lines.
The CNN backbone is a MobileNetV2 \cite{sandler2018mobilenetv2} model with $7$ identical blocks each with a filter dimension of $64$ and an expansion ratio of $8$. The following strided convolution has $128$ filters.
The transformer encoder stack consists of $8$ encoder layers, with hidden size of $256$ and $4$ heads for each layer. 
The inner layers of FFNs in transformer encoders have a hidden size of $512$.
We use a single layer transformer decoder with hidden size $256$ and only $1$ attention head.
The character classification head is trained to recognize case-sensitive Latin characters, digits and printable punctuation symbols (see \textit{supplementary material Sect. B}).
We set $\lambda_4=0.05$ for the bounding box loss.
L2C2W is trained on $16$ TPUv3 cores for $200K$ iterations with a batch size of $1024$ and the same optimizer setup as UDP.
The training data consists of SynthText \cite{gupta2016synthetic}, Synth90K \cite{jaderberg2014synthetic}, HierText \cite{long2022towards}, ICDAR 2015 \cite{karatzas2015icdar}, Total-Text \cite{ch2017total}, CTW1500 \cite{yuliang2017detecting}, and an internal dataset of $1M$ synthetic text lines, with a sampling ratio of $[0.25, 0.20, 0.25, 0.0005, 0.001, 0.001, 0.25]$.
From the full-image datasets \cite{karatzas2015icdar,gupta2016synthetic,ch2017total,yuliang2017detecting,long2022towards}, we use the ground-truth text polygon to crop and rectify text.
SynthText and HierText provide word and line-level annotations. 
The internal synthetic dataset generation process utilizes a similar method to Synth90K \cite{jaderberg2014synthetic} but mainly contains text lines instead of single words\footnote{See Supp. Sect. C. The dataset will be made publicly available.}.
SynthText and our internal synthetic dataset provide annotations for character-level bounding boxes.

\noindent \textbf{Evaluation Practices} Unless specified, e.g. in ablation studies, we use the same model and weights in all experiments. We do not perform fine-tuning on individual datasets.
During inference, we filter the model's output with a confidence threshold of $0.5$ for the detector and $0.8$ for the recognizer.
We determine these thresholds on the HierText validation set and apply them to all experiments.

\subsection{Results on End-to-End Text Spotting}

\begin{table*}
\scalebox{0.88}{
\begin{tabular}{c|cccccccc|cccccc}
\multirow{3}{*}{Method} & \multicolumn{8}{c|}{ICDAR 2015 Incidental} & \multicolumn{6}{c}{Total-Text} \\
 & \multicolumn{4}{c}{Word-Spotting} & \multicolumn{4}{c|}{End-to-End} & \multicolumn{3}{c}{N} & \multicolumn{3}{c}{Full} \\
 & S & W & G & N & S & W & G & N & P & R & F1 & P & R & F1 \\ \hline
MTSv3$\star$ \cite{liao2020mask} & 83.1 & 79.1 & 75.1  & - & 83.3 & 78.1  & 74.2  & - & - & - & 71.2 & - & - & 78.4 \\
MANGO$\star$ \cite{qiao2021mango} & 85.2 & 81.1 & 74.6  & - & 85.4 & 80.1  & 73.9  & - & - & - & 68.9 & - & - & 78.9 \\
YAMTS$\star$ \cite{krylov2021open} & 86.8 & 82.4 & 76.7  & - & 85.3 & 79.8  & 74 & - & - & - & 71.1 & - & - & 78.4 \\
CharNet \cite{xing2019convolutional} & - & - & - & - & 83.10 & 79.15 & 69.14 & 65.73  & - & - & 69.2 & - & - & - \\
TESTR$\sharp$ \cite{zhang2022text} & - & - & - & - & 85.16 & 79.36 & 73.57 & 65.27  & - & - & 73.3 & - & - & {83.9} \\
Qin et al.$\sharp$ \cite{qin2019towards} & - & - & -  & - & 85.51 & 81.91  & - & 69.94 & - & - & 70.7 & - & - & - \\
TTS$\star$ \cite{kittenplon2022towards} & 85.0 & 81.5 & 77.3  & - & 85.2 & 81.7  & {77.4}  & - & - & - & 75.6 & - & - & {84.4} \\
GLASS$\sharp$ \cite{ronen2022glass} & 86.8 & 82.5 & 78.8  & 71.69* & 84.7 & 80.1  & 76.3  & 70.15* & - & - & 76.6 & - & - & 83 \\ 
UNITS $\star$ \cite{kil2023towards} & 88.1 & 84.9 & 80.7  & \textbf{78.7} & 88.4 & 83.9  & 79.7  & 78.5 & - & - & 77.3 & - & - & 85.0 \\ 
\hline
HTS$\star\sharp$ (ours) & \textbf{89.55} & \textbf{85.72} & \textbf{81.23} & {78.62} & \textbf{89.38} & \textbf{84.61} & \textbf{80.69} & \textbf{78.81}  & 80.41 & 75.92 & \textbf{78.10} & 90.12 & 80.74 & \textbf{85.17} 
\end{tabular}
}
\caption{\textbf{Results for ICDAR 2015 and Total-Text.}
‘S’,
‘W’, ‘G’ and ‘N’  refer to \textit{strong}, \textit{weak}, \textit{generic} and \textit{no} lexicons.
‘Full’ for Total-Text means all test set words, and is equivalent to the \textit{weak} setting in ICDAR 2015.
‘-’ means scores are not reported by the papers.
‘*’ means scores are obtained from the open-source code and weights.
$\star$ means models are not fine-tuned on individual datasets.
$\sharp$ means models recognize all symbol classes, including case-sensitive characters and punctuation symbols.
}
\vspace{-3mm}
\label{tab:main}
\end{table*}

\subsubsection{Comparison with State-of-the-Art Results}

We evaluate the proposed HTS method on ICDAR 2015 Incidental \cite{karatzas2015icdar} and Total-Text \cite{ch2017total}, the most popular benchmarks for straight and curved text respectively, and compare our results with current state-of-the-art.
The evaluation of ICDAR 2015 is case-insensitive and includes several heuristics with regard to punctuation symbols and text length.
In the \textit{End-to-End} mode, if a ground-truth text starts with or ends with punctuation, it is considered a true positive match whether or not the prediction includes those punctuation symbols.
In the \textit{Word-Spotting} mode, both ground-truth and predictions are normalized by: 
(1) removing the \textit{'s} and \textit{'S} suffix; 
(2) removing dash (\textit{'-'}) prefix and suffix; 
(3) remove other punctuation symbols; 
(4) only keep normalized words that are at least $3$ letters long.
The Total-Text dataset does not provide an evaluation script for text spotting, and previous works evaluate their results using a script\footnote{\raggedright\url{https://github.com/MhLiao/MaskTextSpotterV3/tree/master/evaluation/totaltext/e2e}} adapted from ICDAR 2015's.
This script inherits the aforementioned heuristics but computes IoU based on polygons as opposed to rotated bounding boxes.
Additionally, both datasets are evaluated with the help of lexicon lists of different levels of perplexity\footnote{\url{https://rrc.cvc.uab.es/?ch=4&com=tasks}}.

To adapt to these heuristics, we transform the output of HTS by: 
(1) convert all characters to lower cases; 
(2) remove all non-alphanumeric symbols; 
(3) remove a detection if it consists of only punctuation symbols; 
(4) use edit distance to pick the best-match when lexicons are used. 

Table~\ref{tab:main} summarizes our evaluation results.
We mark methods with different labels based on: a) the ability to recognize case-sensitive or case-insensitive characters and b) whether the underlying models are fine-tuned on the target dataset.
On ICDAR 2015, our proposed HTS surpasses the recent state-of-the-art UNITS \cite{kil2023towards} considerably and beats previous ones significantly without fine-tuning.
In the \textit{Word-Spotting} mode, HTS has a large margin of \textcolor{teal}{\textbf{+1.45}} / \textcolor{teal}{\textbf{+0.82}} / \textcolor{teal}{\textbf{+0.53}} on S/W/G lexicons and almost matches UNITS on non-lexicon. 
In the \textit{End-to-End} mode, HTS also achieves considerable margins on all lexicon settings.
Note that, the strongest competitor UNITS uses additional training data from TextOCR \cite{Singh_2021_CVPR} while we don't, demonstraing the advantage of our method.

On Total-Text, we surpass all current state-of-the-art in both settings. 
Some of these prior arts \cite{ronen2022glass,kittenplon2022towards,zhang2022text} fine-tune their models on Total-Text which boosts the performance on this target dataset at the cost of dropping performance on others.
Also note that, some prior arts \cite{liao2020mask,qiao2021mango,krylov2021open,kittenplon2022towards} limit recognition to case-insensitive letters and no punctuation symbols, while ours operate in a case-sensitive mode, a more difficult but more important one.
This is not reflected in the scores due to the text normalization rules in the evaluation protocol.

\subsubsection{Comparison based on HierText's Eval}

We also compare the proposed HTS method with others under the evaluation protocol\footnote{\url{https://github.com/google-research-datasets/hiertext/blob/main/eval.py}} of HierText \cite{long2022towards}.
The HierText protocol directly compares predictions against ground-truth, without normalizing letter cases, punctuation symbols, or filtering based on text lengths.
It does not have lexicon modes either.
Compared with the ICDAR 2015 protocol, it provides a more strict and comprehensive comparison since letter cases, punctuation symbols, and text of different lengths are all important in real-scenario applications.

\begin{table}
\resizebox{\columnwidth}{!}{
\begin{tabular}{l|rrl|rrl|rrr}
\multicolumn{1}{c|}{\multirow{2}{*}{Method}} & \multicolumn{3}{c|}{ICDAR 2015} & \multicolumn{3}{c|}{Total-Text} & \multicolumn{3}{c}{HierText test} \\
\multicolumn{1}{c|}{} & \multicolumn{1}{c}{P} & \multicolumn{1}{c}{R} & \multicolumn{1}{c|}{F1} & \multicolumn{1}{c}{P} & \multicolumn{1}{c}{R} & \multicolumn{1}{c|}{F1} & \multicolumn{1}{c}{P} & \multicolumn{1}{c}{R} & \multicolumn{1}{c}{F1} \\ \hline
TESTR & 65.52 & 68.08 & 66.78 & 59.40 & {68.33} & 63.55 & 65.05 & 44.89 & 53.12 \\
MTSv3 & 63.89 & 58.88 & 61.28 & 64.13 & 62.85 & 63.48 & 66.61 & 41.29 & 50.98 \\
GLASS & 74.11 & 63.08 & 68.15 & 68.54 & 60.12 & 64.05 & 73.84 & 57.20 & 64.47 \\ \hline
\multicolumn{1}{c|}{\begin{tabular}[c]{@{}c@{}}HTS \\ (ours)\end{tabular}} & \textbf{81.87} & \textbf{68.41} & \textbf{74.53} & \textbf{75.65} & \textbf{69.43} & \textbf{72.40} & \textbf{86.71} & \textbf{68.48} & \multicolumn{1}{l}{\textbf{76.52}}
\end{tabular}
}
\caption{
Results under the evaluation protocol of HierText.
}
\vspace{-2mm}
\label{tab:hiertext}
\end{table}

For a more fair comparison, we \textbf{re-train} several previous state-of-the-art methods that have opensourced code by the time of this work, including MTSv3 \cite{liao2020mask}, TESTR \cite{zhang2022text} and GLASS \cite{ronen2022glass}, using their open-source codes. 
We use the same combination of HierText \cite{long2022towards}, Total-Text \cite{ch2017total}, CTW1500 \cite{yuliang2017detecting}, SynthText \cite{gupta2016synthetic}, and ICDAR 2015 \cite{karatzas2015icdar} as training data and evaluate on HierText \cite{long2022towards}, Total-Text \cite{ch2017total}, and ICDAR 2015 \cite{karatzas2015icdar}.
We obtain results on HierText test set using the online platform\footnote{\url{https://rrc.cvc.uab.es/?ch=18}} since the test set annotation is not released.
We are also the first to report results on the HierText test set \cite{long2022towards}.
Results are summarized in Table~\ref{tab:hiertext}.

Our HTS achieves significant advantage over these baselines by a large margin on all datasets, proving the effectiveness of our method across straight and curved text, and sparse and dense text.
For ICDAR 2015 and Total-Text, the performance gap with Tab. \ref{tab:main} highlights the impact of text normalization and the use of lexicon lists, and that with such heuristics in evaluation we tend to overestimate the progress of text spotting method's accuracy.
HierText is a new dataset that is characterized by its high word density of more than $100$ words per image, a variety of image domains, a diversity in text sizes and locations and an abundance in text lines that have plenty of punctuation symbols.
The recall rate is lower than $45\%$ for TESTR and MTSv3, and lower than $60\%$ for GLASS, while our proposed HTS can recall more than $68\%$ of words.
That indicates that the word-centric design of most existing text spotting models is not optimal for  natural images with high text density.

\subsection{Results on Geometric Layout Analysis}
\begin{figure}[t]
  \includegraphics[width=1.\linewidth]{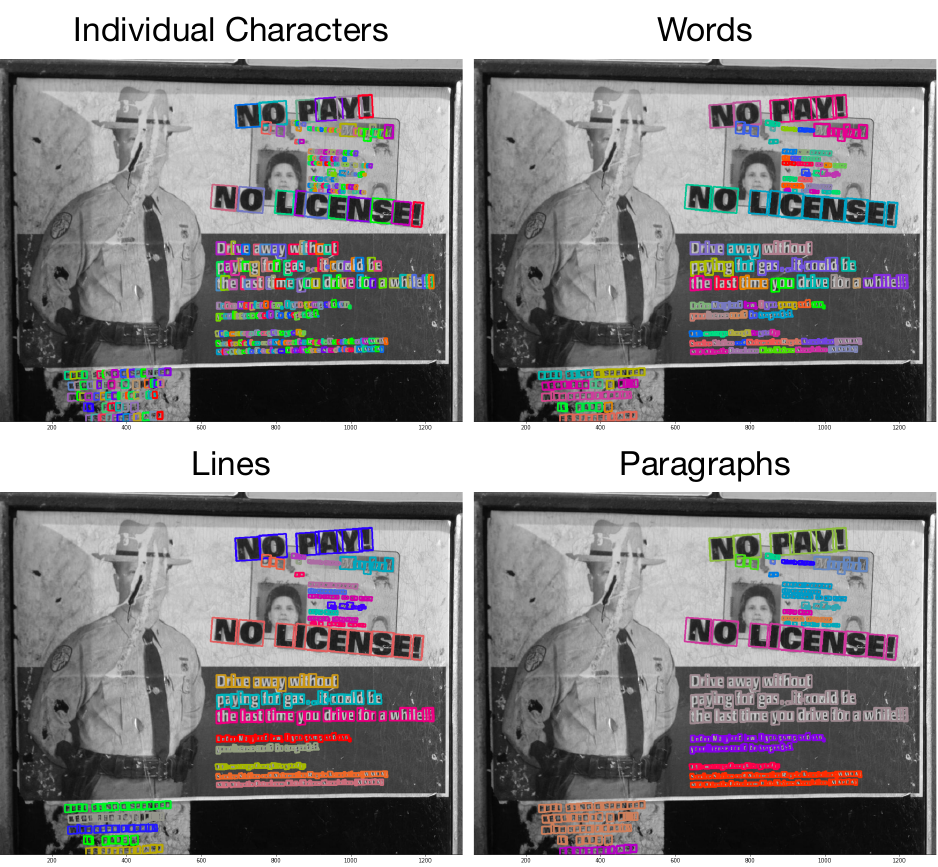}

  \caption{
  \textbf{Qualitative results for layout analysis.} 
  We draw character bounding boxes with different colors to indicate layout at different levels.
  }
  \label{fig:demo}
\end{figure}

\begin{table}[]
\resizebox{\columnwidth}{!}{%
\begin{tabular}{|c|ccccc|ccccc|}
\hline
\multirow{2}{*}{Method} & \multicolumn{5}{c|}{Line Grouping} & \multicolumn{5}{c|}{Paragraph Grouping} \\ \cline{2-11} 
 & \multicolumn{1}{c|}{P} & \multicolumn{1}{c|}{R} & \multicolumn{1}{c|}{F1} & \multicolumn{1}{c|}{T} & PQ & \multicolumn{1}{c|}{P} & \multicolumn{1}{c|}{R} & \multicolumn{1}{c|}{F1} & \multicolumn{1}{c|}{T} & PQ \\ \hline
\begin{tabular}[c]{@{}c@{}}Unified\\ Detector \cite{long2022towards}\end{tabular} & \multicolumn{1}{c|}{79.64} & \multicolumn{1}{c|}{80.19} & \multicolumn{1}{c|}{79.91} & \multicolumn{1}{c|}{77.87} & 62.23 & \multicolumn{1}{c|}{\textbf{76.04}} & \multicolumn{1}{c|}{62.45} & \multicolumn{1}{c|}{68.58} & \multicolumn{1}{c|}{78.17} & 53.60 \\ \hline
HTS (ours) & \multicolumn{1}{c|}{\textbf{82.71}} & \multicolumn{1}{c|}{\textbf{82.03}} & \multicolumn{1}{c|}{\textbf{82.37}} & \multicolumn{1}{c|}{\textbf{80.51}} & \textbf{66.31} & \multicolumn{1}{c|}{{75.26}} & \multicolumn{1}{c|}{\textbf{75.98}} & \multicolumn{1}{c|}{\textbf{75.62}} & \multicolumn{1}{c|}{\textbf{79.67}} & \textbf{60.25} \\ \hline
\end{tabular}
}
\caption{
\textbf{Results of geometric layout analysis on HierText test set.}
\textit{Panoptic Quality}, equals to the product of \textit{F1} and \textit{Tightness}.
}
\label{tab:layout}
\end{table}

The proposed HTS is able to estimate the text's layout structure in images, as shown in Fig. \ref{fig:demo}.
We further evaluate our model on HierText on the geometric layout analysis task, and summarize the results in Table~\ref{tab:layout}.
HTS achieves betters scores in the PQ metric on both line (\textcolor{teal}{\textbf{+4.08}}) and paragraph grouping (\textcolor{teal}{\textbf{+6.65}}) compared to Unified Detector. 
Most notably, line and paragraph predictions are formed as union masks of underlying character boxes.
This indicates that our character localization, as well as the word box estimation based on it, are accurate.

\subsection{Ablation Studies}
\label{sec:ablation}
To better understand the effectiveness of our design choices,
we conduct ablation studies and summarize the results in Tab. \ref{tab:ablation}. 
Different from the previous sections, here we use $1024\times1024$ as input image resolution.

\begin{figure}[t]
\begin{center}
  \includegraphics[width=1.\linewidth]{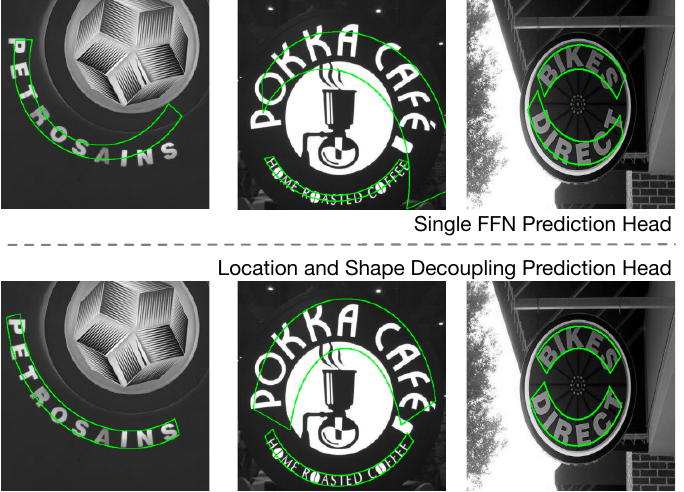}

  \caption{
  \textbf{Qualitative comparison} between Single FFN prediction head (Top) and the proposed \textit{Location and Shape Decoupling Module (LSDM)} prediction head (Bottom) for Bezier curve polygons.
  The original images are turned to gray for clearer views.
  }
  \label{fig:lsd}
  \end{center}
\end{figure}

\begin{table}[]
\resizebox{\columnwidth}{!}{%
\begin{tabular}{c|cccccc}
\hline
\multirow{2}{*}{Abaltion} & \multicolumn{3}{c}{Total-Text} & \multicolumn{3}{c}{HierText Val} \\
 & P & R & F1 & P & R & F1 \\ \hline
Full & \textbf{72.33} & \textbf{62.95} & \textbf{67.31} & {86.54} & \textbf{67.03} & \textbf{75.55} \\ \hline
w/o LSDM & 68.79 & 56.12 & 61.81 & \textbf{86.75} & 60.97 & 71.61 \\
Mask-based detection & 64.52 & 54.39 & 59.02 & 84.87 & 63.52 & 72.66 \\
Word-level detection & 71.54 & 58.61 & 64.43 & 80.71 & 51.94 & 63.21 \\ 
MaX-DeepLab & 70.16 & 59.32 & 64.28 & 87.66 & 65.07 & 74.69 \\ 
\hline
\end{tabular}%
}
\caption{
Ablation results as evaluated on the text spotting task.
}
\label{tab:ablation}
\end{table}

\noindent\textbf{LSDM} We replace our LSDM prediction head with a single FFN branch prediction head to produce the Global Bezier directly, and remove the $2$ loss items on AABB.
We use $\lambda_5=3$ for the weight of this loss, which is larger than $\lambda_{1-3}$ to compensate for the difference of loss scales.
The model is trained on the same combination of HierText and CTW1500.
As shown in Tab. \ref{tab:ablation}, removal of LSDM results in a sharp drop of text spotting performances on both Total-Text (\textcolor{red}{-5.50}) and HierText validation set (\textcolor{red}{-3.94}).
Fig. \ref{fig:lsd} further demonstrates that LSDM is important for the learning of text shapes.
Without LSDM, predicted location is only roughly correct but its shape is inaccurate.
This shows that training Bezier prediction head with L1 loss on diverse datasets could be dominated by the location learning and thus shape prediction fails.
The proposed LSDM, on the other hand, can solve this issue by separating and balancing the learning of location and shape.

\noindent\textbf{Mask v.s. Polygon}
One main difference between our UDP with Unified Detector \cite{long2022towards} is that UDP produces polygons as output as opposed to masks.
In this ablation study, we use the mask outputs as detections and find minimum-area rotated bounding boxes instead of using the predicted Bezier polygons.
This results in a significant drop in Total-Text (\textcolor{red}{-8.29}) and a less severe drop in HierText (\textcolor{red}{-2.89}).
Mask representation is unsuitable for curved text spotting since it is non-trivial to crop and rectify with masks.
Note that HierText consists mostly of straight text and is thus less affected.

\noindent\textbf{Word Based v.s. Line Based OCR}
We train HTS on HierText and Total-Text for the word spotting task, as opposed to line-level.
The line-based model is better than the word based model on both Total-Text (\textcolor{teal}{+2.88}), a sparse text dataset, and HierText (\textcolor{teal}{+12.34}), a dense text dataset (Tab. \ref{tab:ablation}).
The recall rate on HierText drops by (\textcolor{red}{-15.09}) if the model detects words instead of lines.
It is consistent with Tab. \ref{tab:hiertext}, where word-based current arts have much lower scores on the dense HierText dataset.

\noindent\textbf{Choice of Backbone} We train two versions of HTS, one with MaX-DeepLab as the backbone and the other with KMaX-DeepLab as backbone. 
HTS with KMaX-DeepLab achieves \textcolor{teal}{\textbf{+3.03}} / \textcolor{teal}{\textbf{+0.86}} better F1 scores on Total-Text and HierText respectively, demonstrating the advantage of KMaX-DeepLab, a follow-up model of MaX-DeepLab. 
In addition to improved accuracy, the KMaX-DeepLab version of HTS can run at a much higher speed, with a $5.6$ FPS on average on HierText, while the MaX-DeepLab version is $1.2$ FPS, when measured on A100.
Adopting KMaX-DeepLab benefits both accuracy and latency.

\subsection{Limitations}
\noindent \textbf{Latency} On a  A100 GPU, our method runs at $7.8$ FPS on ICDAR 2015 and Total-Text, and $5.6$ FPS on HierText which is an order of magnitude more dense, while  TESTR \cite{zhang2022text} runs at $10.2$ FPS on HierText.
We believe a faster backbone for UDP can help.
Sharing features for UDP and L2C2W i.e. making it end-to-end trainable will also save computations.

\noindent \textbf{Line labels} The training of UDP requires line level annotations which are  limited in few public datasets.
However, it is relatively low-cost to annotate line grouping on top of existing word-level polygons.
Line grouping of ground-truth words can also be accurately estimated using heuristics based on word size, location, and orientation.

\noindent \textbf{Character Localization} Benchmark datasets used in this work do not provide character-level labels so we are unable to evaluate the accuracy of our character localization.
We can only use word level text spotting and layout analysis results as a proxy as it highly depends on character localization quality.

\section{Conclusion}
In this paper, we propose the first Hierarchical Text Spotter (HTS) for the joint task of text spotting and layout analysis.
HTS achieves new state-of-the-art performance on multiple word-level text spotting benchmarks as well as a geometric layout analysis task.

{\small
\bibliographystyle{ieee_fullname}
\bibliography{egbib}
}

\end{document}